# Gotta Learn Fast:
# A New Benchmark for Generalization in RL

Alex Nichol, Vicki Pfau, Christopher Hesse, Oleg Klimov, John Schulman

OpenAI

{*alex, vickipfau, csh, oleg, joschu*}*@openai.com*

## Abstract

In this report, we present a new reinforcement learning (RL) benchmark based on the *Sonic the Hedgehog$^{TM}$* video game franchise. This benchmark is intended to measure the performance of transfer learning and few-shot learning algorithms in the RL domain. We also present and evaluate some baseline algorithms on the new benchmark.

# 1 Motivation

In the past few years, it has become clear that deep reinforcement learning can solve difficult, high-dimensional problems when given a good reward function and unlimited time to interact with the environment. However, while this kind of learning is a key aspect of intelligence, it is not the only one. Ideally, intelligent agents would also be able to generalize between tasks, using prior experience to pick up new skills more quickly. In this report, we introduce a new benchmark that we designed to make it easier for researchers to develop and test RL algorithms with this kind of capability.

Most popular RL benchmarks such as the ALE [1] are not ideal for testing generalization between similar tasks. As a result, RL research tends to "train on the test set", boasting an algorithm's final performance on the same environment(s) it was trained on. For the field to advance towards algorithms with better generalization properties, we need RL benchmarks with proper splits between "train" and "test" environments, similar to supervised learning datasets. Our benchmark has such a split, making it ideal for measuring cross-task generalization.

One interesting application of cross-task generalization is few-shot learning. Recently, supervised few-shot learning algorithms have improved by leaps and bounds [2]–[4]. This progress has hinged on the availability of good meta-learning datasets such as Omniglot [5] and Mini-ImageNet [6]. Thus, if we want better few-shot RL algorithms, it makes sense to construct a similar kind of dataset for RL. Our benchmark is designed to be a meta-learning dataset, consisting of many similar tasks sampled from a single task distribution. Thus, it is a suitable test bed for few-shot RL algorithms.

Beyond few-shot learning, there are many other applications of cross-task generalization that require the right kind of benchmark. For example, you might want an RL algorithm to learn how to explore in new environments. Our benchmark poses a fairly challenging exploration problem, and the train/test split presents a unique opportunity to learn how to explore on some levels and transfer this ability to other levels.

# 2    Related Work

Our Gym Retro project, as detailed in Section 3.1, is related to both the Retro Learning Environment (RLE) [7] and the Arcade Learning Environment (ALE) [1]. Unlike these projects, however, Gym Retro aims to be flexible and easy to extend, making it straightforward to create a huge number of RL environments.

Our benchmark is related to other meta-learning datasets like Omniglot [5] and Mini-ImageNet [6]. In particular, our benchmark is intended to serve the same purpose for RL as datasets like Omniglot serve for supervised learning.

Our baselines in Section 4 explore the ability of RL algorithms to transfer between video game environments. Several prior works have reported positive transfer results in the video game setting:

- Parisotto et al. [8] observed that pre-training on certain Atari games could increase a network's learning speed on other Atari games.

- Rusu et al. [9] proposed a new architecture for transfer learning called progressive networks, and showed that it could boost learning speed across a variety of previously unseen Atari games.

- Pathak et al. [10] found that an exploratory agent trained on one level of *Super Mario Bros.* could be used to boost performance on two other levels.

- Fernando et al. [11] found that their PathNet algorithm increased learning speed on average when transferring from one Atari game to another.

- Higgins et al. [12] used an unsupervised vision objective to produce robust features for a policy, and found that this policy was able to transfer to previously unseen vision tasks in DeepMind Lab [13] and MuJoCo [14].

In previous literature on transfer learning in RL, there are two common evaluation techniques: evaluation on synthetic tasks, and evaluation on the ALE. The former evaluation technique is rather ad hoc and makes it hard to compare different algorithms, while the latter typically reveals fairly small gains in sample complexity. One problem with the ALE in particular is that all the games are quite different, meaning that it may not be possible to get large improvements from transfer learning.

Ideally, further research in transfer learning would be able to leverage a standardized benchmark that is difficult like the ALE but rich with similar environments like well-crafted synthetic tasks. We designed our proposed benchmark to satisfy both criteria.

# 3    The Sonic Benchmark

This section describes the Sonic benchmark in detail. Each subsection focuses on a different aspect of the benchmark, ranging from technical details to high-level design features.

## 3.1 Gym Retro

Underlying the Sonic benchmark is Gym Retro, a project aimed at creating RL environments from various emulated video games. At the core of Gym Retro is the gym-retro Python package, which exposes emulated games as Gym [15] environments. Like RLE [7], gym-retro uses the libretro API[1] to interface with game emulators, making it very easy to add new emulators to gym-retro.

The gym-retro package includes a dataset of games. Each game in the dataset consists of a *ROM*, one or more *save states*, one or more *scenarios*, and a *data file*. Here are high-level descriptions of each of these components:

- ROM – the data and code that make up a game; loaded by an emulator to play that game.

- Save state – a snapshot of the console's state at some point in the game. For example, a save state could be created for the beginning of each level.

- Data file – a file describing where various pieces of information are stored in console memory. For example, a data file might indicate where the score is located.

- Scenario – a description of done conditions and reward functions. A scenario file can reference fields from the data file.

## 3.2 The Sonic Video Game

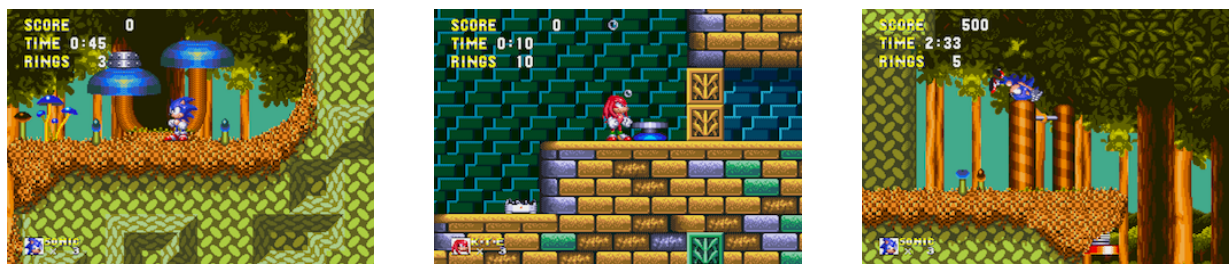

Figure 1: Screenshots from *Sonic 3 & Knuckles*. **Left:** a situation where the player can be shot into the air by utilizing an object with lever-like dynamics (Mushroom Hill Zone, Act 2). **Middle:** a door that opens when the player jumps on a button (Hydrocity Zone, Act 1). **Right:** a swing that the player must jump from at exactly the right time to reach a high platform (Mushroom Hill Zone, Act 2).

In this benchmark, we use three similar games: *Sonic The Hedgehog$^{TM}$*, *Sonic The Hedgehog$^{TM}$2*, and *Sonic 3 & Knuckles*. All of these games have very similar rules and controls, although there are subtle differences between them (e.g. *Sonic 3 & Knuckles* includes some extra controls and characters). We use multiple games to get as many environments for our dataset as possible.

---

[1] https://www.libretro.com/index.php/api

Each Sonic game is divided up into *zones*, and each zone is further divided up into *acts*. While the rules and overarching objective remain the same throughout the entire game, each zone has a unique set of textures and objects. Different acts within a zone tend to share these textures and objects, but differ in spatial layout. We will refer to a (`ROM, zone, act`) tuple as a "level".

The Sonic games provide a rich set of challenges for the player. For example, some zones include platforms that the player must jump on in order to open doors. Other zones require the player to first jump on a lever to send a projectile into the air, then wait for the projectile to fall back on the lever to send the player over some sort of obstacle. One zone even has a swing that the player must jump off of at a precise time in order to launch Sonic up to a higher platform. Examples of these challenges are presented in Figure 1.

### 3.3 Games and Levels

Our benchmark consists of a total of 58 save states taken from three different games, where each of these save states has the player at the beginning of a different level. A number of acts from the original games were not used because they contained only boss fights or because they were not compatible with our reward function.

We split the test set by randomly choosing zones with more than one act and then randomly choosing an act from each selected zone. In this setup, the test set contains mostly objects and textures present in the training set, but with different layouts.

The test levels are listed in the following table:

| ROM | Zone | Act |
|---|---|---|
| Sonic The Hedgehog | SpringYardZone | 1 |
| Sonic The Hedgehog | GreenHillZone | 2 |
| Sonic The Hedgehog | StarLightZone | 3 |
| Sonic The Hedgehog | ScrapBrainZone | 1 |
| Sonic The Hedgehog 2 | MetropolisZone | 3 |
| Sonic The Hedgehog 2 | HillTopZone | 2 |
| Sonic The Hedgehog 2 | CasinoNightZone | 2 |
| Sonic 3 & Knuckles | LavaReefZone | 1 |
| Sonic 3 & Knuckles | FlyingBatteryZone | 2 |
| Sonic 3 & Knuckles | HydrocityZone | 1 |
| Sonic 3 & Knuckles | AngelIslandZone | 2 |

### 3.4 Frame Skip

The `step()` method on raw gym-retro environments progresses the game by roughly $\frac{1}{60}^{th}$ of a second. However, following common practice for ALE environments, we require the use of a frame skip [16] of 4. Thus, from here on out, we will use *timesteps* as the main unit of measuring in-game time. With a frame skip of 4, a timestep represents roughly $\frac{1}{15}^{th}$ of a second. We believe that this is more than enough temporal resolution to play Sonic well.

Moreover, since deterministic environments are often susceptible to trivial scripted solutions [17], we require the use of a stochastic "sticky frame skip". Sticky frame skip adds a small amount of randomness to the actions taken by the agent; it does not directly alter observations or rewards.

Like standard frame skip, sticky frame skip applies $n$ actions over $4n$ frames. However, for each action, we delay it by one frame with probability 0.25, applying the previous action for that frame instead. The following diagram shows an example of an action sequence with sticky frame skip:

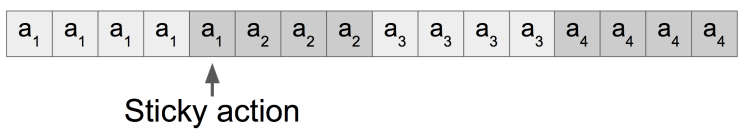

Sticky action

## 3.5 Episode Boundaries

Experience in the game is divided up into episodes, which roughly correspond to lives. At the end of each episode, the environment is reset to its original save state. Episodes can end on three conditions:

- The player completes a level successfully. In this benchmark, completing a level corresponds to passing a certain horizontal offset within the level.

- The player loses a life.

- 4500 timesteps have elapsed in the current episode. This amounts to roughly 5 minutes of in-game time.

The environment should only be reset if one of the aforementioned done conditions is met. Agents should not use special APIs to tell the environment to start a new episode early.

Note that our benchmark omits the boss fights that often take place at the end of a level. For levels with boss fights, our done condition is defined as a horizontal offset that the agent must reach before the boss fight. Although boss fights could be an interesting problem to solve, they are fairly different from the rest of the game. Thus, we chose not to include them so that we could focus more on exploration, navigation, and speed.

## 3.6 Observations

A gym-retro environment produces an observation at the beginning of every timestep. This observation is always a 24-bit RGB image, but the dimensions vary by game. For Sonic, the screen images are 320 pixels wide and 224 pixels tall.

## 3.7 Actions

At every timestep, an agent produces an action representing a combination of buttons on the game console. Actions are encoded as binary vectors, where 1 means "pressed" and 0 means "not pressed". For Sega Genesis games, the action space contains the following buttons: `B, A, MODE, START, UP, DOWN, LEFT, RIGHT, C, Y, X, Z`.

A small subset of all possible button combinations makes sense in Sonic. In fact, there are only eight essential button combinations:

$$\{\{\}, \{\text{LEFT}\}, \{\text{RIGHT}\}, \{\text{LEFT, DOWN}\},$$
$$\{\text{RIGHT, DOWN}\}, \{\text{DOWN}\}, \{\text{DOWN, B}\}, \{\text{B}\}\}$$

The `UP` button is also useful on occasion, but for the most part it can be ignored.

## 3.8 Rewards

During an episode, agents are rewarded such that the cumulative reward at any point in time is proportional to the horizontal offset from the player's initial position. Thus, going right always yields a positive reward, while going left always yields a negative reward. This reward function is consistent with our done condition, which is based on the horizontal offset in the level.

The reward consists of two components: a horizontal offset, and a completion bonus. The horizontal offset reward is normalized per level so that an agent's total reward will be 9000 if it reaches the predefined horizontal offset that marks the end of the level. This way, it is easy to compare scores across levels of varying length. The completion bonus is 1000 for reaching the end of the level instantly, and drops linearly to zero at 4500 timesteps. This way, agents are encouraged to finish levels as fast as possible[2].

Since the reward function is dense, RL algorithms like PPO [18] and DQN [16] can easily make progress on new levels. However, the immediate rewards can be deceptive; it is often necessary to go backwards for prolonged amounts of time (Figure 2). In our RL baselines, we use reward preprocessing so that our agents are not punished for going backwards. Note, however, that the preprocessed reward still gives no information about when or how an agent should go backwards.

## 3.9 Evaluation

In general, all benchmarks must provide some kind of performance metric. For Sonic, this metric takes the form of a "mean score" as measured across all the levels in the test set. Here are the general steps for evaluating an algorithm on Sonic:

1. At training time, use the training set as much or as little as you like.

2. At test time, play each test level for 1 million timesteps. Play each test level separately; do not allow information to flow between test levels. Multiple copies of each environment may be used (as is done in algorithms like A3C [19]).

3. For each 1 million timestep evaluation, average the total reward per episode across all episodes. This gives a per-level mean score.

4. Average the mean scores for all the test levels, giving an aggregate metric of performance.

---

[2]In practice, RL agents may not be able to leverage a bonus at the end of an episode due to a discount factor.

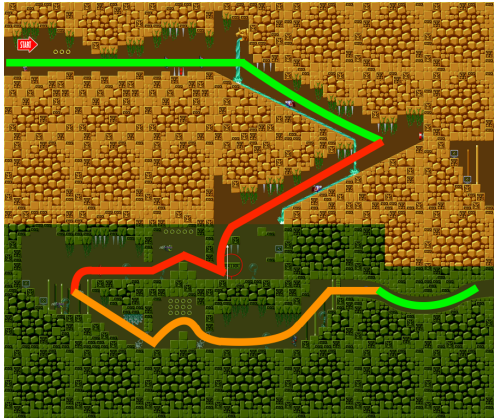

Figure 2: A trace of a successful path through the first part of *Labyrinth Zone, Act 2* in *Sonic The Hedgehog$^{TM}$*. In the initial green segment, the agent is moving rightwards, getting positive reward. In the red segment, the agent must move to the left, getting negative reward. During the orange segment, the agent is once again moving right, but its cumulative reward is still not as high as it was after the initial green segment. In the final green segment, the agent is finally improving its cumulative reward past the initial green segment. For an average player, it takes 20 to 30 seconds to get through the red and orange segments.

The most important aspect of this procedure is the timestep limit for each test level. In the infinite-timestep regime, there is no strong reason to believe that meta-learning or transfer learning is necessary. However, in the limited-timestep regime, transfer learning may be necessary to achieve good performance quickly.

We aim for this version of the Sonic benchmark to be easier than zero-shot learning but harder than $\infty$-shot learning. 1 million timesteps was chosen as the timestep limit because modern RL algorithms can make some progress in this amount of time.

## 4  Baselines

In this section, we present several baseline learning algorithms and discuss their performance on the benchmark. Our baselines include human players, several methods that do not make use of the training set, and a simple transfer learning approach consisting of joint training followed by fine tuning. Table 1 gives the aggregate scores for each of the baselines, and Figure 3 compares the baselines' aggregate learning curves.

### 4.1  Humans

For the human baseline, we had four test subjects play each test level for one hour. Before seeing the test levels, each subject had two hours to practice on the training levels. Table 7 in Appendix C shows average human scores over the course of an hour.

Table 1: Aggregate test scores for each of the baseline algorithms.

| Algorithm | Score |
|---:|:---|
| Rainbow | $2748.6 \pm 102.2$ |
| JERK | $1904.0 \pm 21.9$ |
| PPO | $1488.8 \pm 42.8$ |
| PPO (joint) | $3127.9 \pm 116.9$ |
| Rainbow (joint) | $2969.2 \pm 170.2$ |
| Human | $7438.2 \pm 624.2$ |

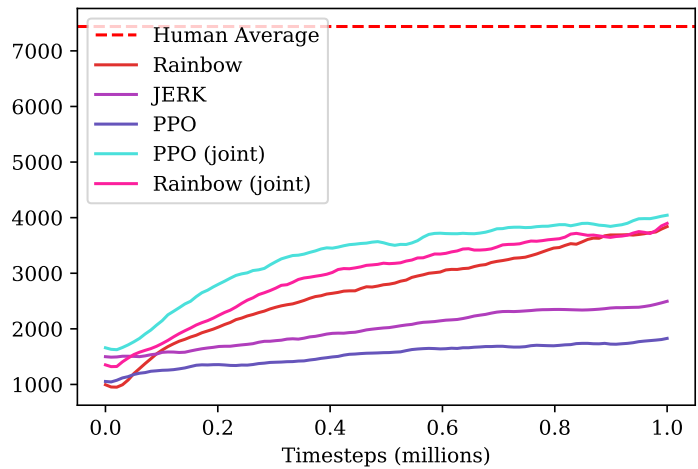

Figure 3: The mean learning curves for all the baselines across all the test levels. Every curve is an average over three runs. The y-axis represents instantaneous score, not average over training.

## 4.2 Rainbow

Deep Q-learning (DQN) [16] is a popular class of algorithms for reinforcement learning in high-dimensional environments like video games. We use a specific variant of DQN, namely Rainbow [20], which performs particularly well on the ALE.

We retain the architecture and most of the hyper-parameters from [20], with a few small changes. First, we set $V_{max} = 200$ to account for Sonic's reward scale. Second, we use a replay buffer size of 0.5M instead of 1M to lower the algorithm's memory consumption. Third, we do not use hyper-parameter schedules; rather, we simply use the initial values of the schedules from [20].

Since DQN tends to work best with a small, discrete action space, we use an action space containing seven actions:

$$\{\{\text{LEFT}\}, \{\text{RIGHT}\}, \{\text{LEFT, DOWN}\}, \{\text{RIGHT, DOWN}\} \\ \{\text{DOWN}\}, \{\text{DOWN, B}\}, \{\text{B}\}\}$$

We use an environment wrapper that rewards the agent based on deltas in the maximum x-position. This way, the agent is rewarded for getting further than it has been before (in the current episode), but it is not punished for backtracking in the level. This reward preprocessing gives a sizable performance boost.

Table 2 in Appendix C shows Rainbow's scores for each test level.

### 4.3 JERK: A Scripted Approach

In this section, we present a simple algorithm that achieves high rewards on the benchmark without using any deep learning. This algorithm completely ignores observations and instead looks solely at rewards. We call this algorithm *Just Enough Retained Knowledge* (JERK). We note that JERK is loosely related to The Brute [21], a simple algorithm that finds good trajectories in deterministic Atari environments without leveraging any deep learning.

Algorithm 1 in Appendix A describes JERK in detail. The main idea is to explore using a simple algorithm, then to replay the best action sequences more and more frequently as training progresses. Since the environment is stochastic, it is never clear which action sequence is the best to replay. Thus, each action sequence has a running mean of its rewards.

Table 3 in Appendix C shows JERK's scores for each test level. We note that JERK actually performs better than regular PPO, which is likely due to JERK's perfect memory and its tailored exploration strategy.

### 4.4 PPO

Proximal Policy Optimization (PPO) [18] is a policy gradient algorithm which performs well on the ALE. For this baseline, we run PPO individually on each of the test levels.

For PPO we use the same action and observation spaces as for Rainbow, as well as the same reward preprocessing. For our experiments, we scaled the rewards by a small constant factor in order to bring the advantages to a suitable range for neural networks. This is similar to how we set $V_{max}$ for Rainbow. The CNN architecture is the same as the one used in [18] for Atari.

We use the following hyper-parameters for PPO:

| Hyper-parameter | Value |
|---|---|
| Workers | 1 |
| Horizon | 8192 |
| Epochs | 4 |
| Minibatch size | 8192 |
| Discount ($\gamma$) | 0.99 |
| GAE parameter ($\lambda$) | 0.95 |
| Clipping parameter ($\epsilon$) | 0.2 |
| Entropy coeff. | 0.001 |
| Reward scale | 0.005 |

Table 4 in Appendix C shows PPO's scores for each test level.

## 4.5 Joint PPO

While Section 4.4 evaluates PPO with no meta-learning, this section explores the ability of PPO to transfer from the training levels to the test levels. To do this, we use a simple joint training algorithm[3], wherein we train a policy on all the training levels and then use it as an initialization on the test levels.

During meta-training, we train a single policy to play every level in the training set. Specifically, we run 188 parallel workers, each of which is assigned a level from the training set. At every gradient step, all the workers average their gradients together, ensuring that the policy is trained evenly across the entire training set. This training process requires hundreds of millions of timesteps to converge (see Figure 4), since the policy is being forced to learn a lot more than a single level. Besides the different training setup, we use the same hyper-parameters as for regular PPO.

Once the joint policy has been trained on all the training levels, we fine-tune it on each test level under the standard evaluation rules. In essence, the training set provides an initialization that is plugged in when evaluating on the test set. Aside from the initialization, nothing is changed from the evaluation procedure used for Section 4.4.

Figure 4 shows that, after roughly 50 million timesteps of joint training, further improvement on the training set stops leading to better performance on the test set. This can be thought of as the point where the model starts to overfit. The figure also shows that zero-shot performance does not increase much after the first few million timesteps of joint training.

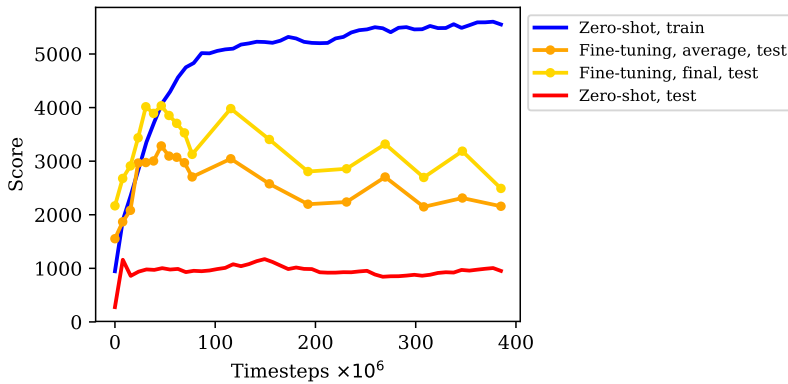

Figure 4: Intermediate performance during the process of joint training a PPO model. The x-axis corresponds to timesteps into the joint training process. The zero-shot curves were densely sampled during training, while the fine-tuning curves were sampled periodically.

Table 5 in Appendix C shows Joint PPO's scores for each test level. Table 9 in Appendix D shows Joint PPO's final scores for each training level. The resulting test performance is superior to that of Rainbow, and is roughly 100% better than that of regular PPO. Thus, it is clear that some kind of useful information is being transferred from the training levels to the test levels.

---

[3]We also tried a version of Reptile [22], but found that it yielded worse results.

### 4.6 Joint Rainbow

Since Rainbow outperforms PPO with no joint training, it is natural to ask if Joint Rainbow analogously outperforms Joint PPO. Surprisingly, our experiments indicate that this is not the case.

To train a single Rainbow model on the entire training set, we use a multi-machine training setup with 32 GPUs. Each GPU corresponds to a single *worker*, where each worker has its own replay buffer and eight environments. The environments are all "joint environments", meaning that they sample a new training level at the beginning of every episode. Each worker runs the algorithm described in Algorithm 2 in Appendix A.

Besides the unusual batch size and distributed worker setup, all the hyper-parameters are kept the same as for the regular Rainbow experiment.

Table 6 in Appendix C shows the performance of fine-tuning on every test level. Table 8 in Appendix D shows the performance of the jointly trained model on every training level.

## 5 Discussion

We have presented a new reinforcement learning benchmark and used it to evaluate several baseline algorithms. Our results leave a lot of room for improvement, especially since our best transfer learning results are not much better than our best results learning from scratch. Also, our results are nowhere close to the maximum achievable score (which, by design, is somewhere between 9000 and 10000).

Now that the benchmark and baseline results have been laid out, there are many directions to take further research. Here are some questions that future research might seek to answer:

- How much can exploration objectives help training performance on the benchmark?
- Can transfer learning be improved using data augmentation?
- Is it possible to improve performance on the test set using a good feature representation learned on the training set (like in Higgins et al. [12])?
- Can different architectures (e.g. Transformers [23] and ResNets [24]) be used to improve training and/or test performance?

While we believe the Sonic benchmark is a step in the right direction, it may not be sufficient for exploring meta-learning, transfer learning, and generalization in RL. Here are some possible problems with this benchmark, which will only be proven or disproven once more work has been done:

- It may be possible to solve a Sonic level in many fewer than 1M timesteps without any transfer learning.
- Sonic-specific hacks may outperform general meta-learning approaches.
- Exploration strategies that work well in Sonic may not generalize beyond Sonic.

- Mastering a Sonic level involves some degree of memorization. Algorithms which are good at few-shot memorization may not be good at other tasks.

# A   Detailed Algorithm Descriptions

---

**Algorithm 1** The JERK algorithm. For our experiments, we set $\beta = 0.25$, $J_n = 4$, $J_p = 0.1$, $R_n = 100$, $L_n = 70$.

---

   **Require:** initial exploitation fraction, $\beta$.
   **Require:** consecutive timesteps for holding the jump button, $J_n$.
   **Require:** probability of triggering a sequence of jumps, $J_p$.
   **Require:** consecutive timesteps to go right, $R_n$.
   **Require:** consecutive timesteps to go left, $L_n$.
   **Require:** evaluation timestep limit, $T_{max}$.
   $S \leftarrow \{\}$, $T \leftarrow 0$.
   **repeat**
      **if** $|S| > 0$ and $RandomUniform(0, 1) < \beta + \frac{T}{T_{max}}$ **then**
         Replay the best trajectory $\tau \in S$. Pad the episode with no-ops as needed.
         Update the mean reward of $\tau$ based on the new episode reward.
         Add the elapsed timesteps to $T$.
      **else**
         **repeat**
            Go right for $R_n$ timesteps, jumping for $J_n$ timesteps at a time with $J_p$ probability.
            **if** cumulative reward did not increase over the past $R_n$ steps **then**
               Go left for $L_n$ timesteps, jumping periodically.
            **end if**
            Add the elapsed timesteps to $T$.
         **until** episode complete
         Find the timestep $t$ from the previous episode with the highest cumulative reward $r$.
         insert $(\tau, r)$ into $S$, where $\tau$ is the action sequence up to timestep $t$.
      **end if**
   **until** $T \geq T_{max}$

---

**Algorithm 2** The joint training procedure for each worker in Joint Rainbow. For our experiments, we set $N = 256$.

    $R \leftarrow$ empty replay buffer.
    $\theta \leftarrow$ initial weights.
    **repeat**
        **for** each environment **do**
            $T \leftarrow$ next state transition.
            add $T$ to $R$.
        **end for**
        $B \leftarrow$ sample $N$ transitions from $R$.
        $L \leftarrow Loss(B)$
        Update the priorities in $R$ according to $L$.
        $G \leftarrow \nabla_\theta L$
        $G_{agg} \leftarrow AllReduce(G)$ (average gradient between workers).
        $\theta \leftarrow Adam(\theta, G_{agg})$
    **until** convergence

## B  Plots for Multiple Seeds

In this section, we present per-algorithm learning curves on the test set. For each algorithm, we run three different random seeds.

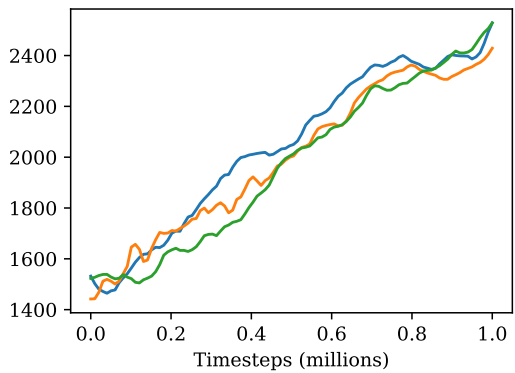

Figure 5: Test learning curves for JERK.

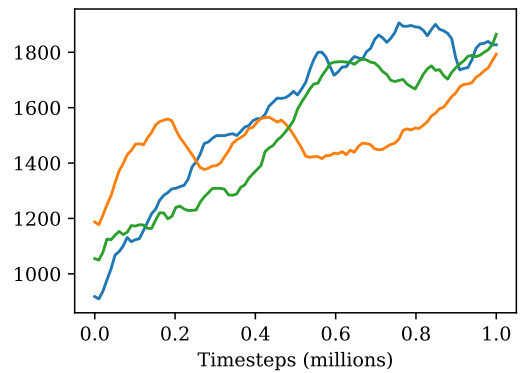

Figure 6: Test learning curves for PPO.

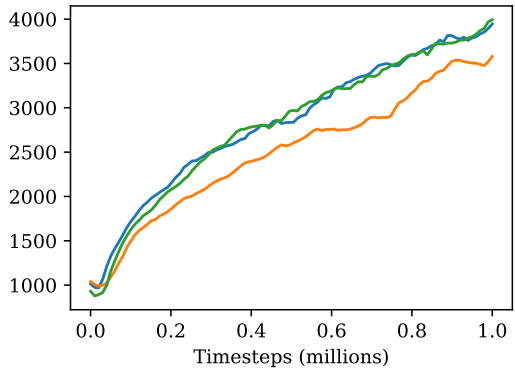

Figure 7: Test learning curves for Rainbow.

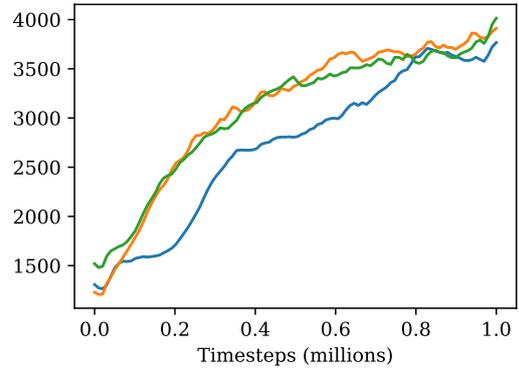

Figure 8: Test learning curves for Joint Rainbow.

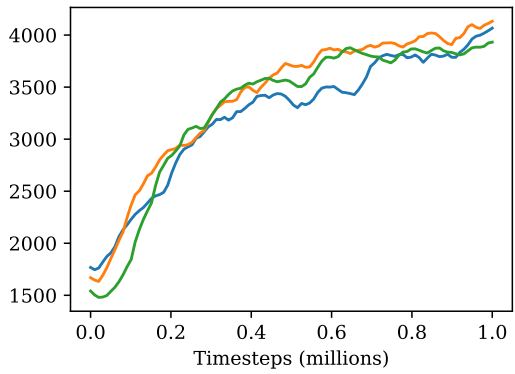

Figure 9: Test learning curves for Joint PPO.

# C  Scores on Test Set

Table 2: Detailed evaluation results for Rainbow.

| State | Score | Final Score |
|---:|---|---|
| AngelIslandZone Act2 | $3576.0 \pm 89.2$ | $5070.1 \pm 433.1$ |
| CasinoNightZone Act2 | $6045.2 \pm 845.4$ | $8607.9 \pm 1022.5$ |
| FlyingBatteryZone Act2 | $1657.5 \pm 10.1$ | $2195.4 \pm 190.8$ |
| GreenHillZone Act2 | $6332.0 \pm 263.5$ | $6817.2 \pm 392.8$ |
| HillTopZone Act2 | $2847.8 \pm 161.9$ | $3432.7 \pm 252.9$ |
| HydrocityZone Act1 | $886.4 \pm 31.4$ | $867.2 \pm 0.0$ |
| LavaReefZone Act1 | $2623.6 \pm 78.0$ | $2908.5 \pm 106.1$ |
| MetropolisZone Act3 | $1178.1 \pm 229.3$ | $2278.8 \pm 280.6$ |
| ScrapBrainZone Act1 | $879.1 \pm 141.0$ | $2050.0 \pm 1089.9$ |
| SpringYardZone Act1 | $1787.6 \pm 136.5$ | $3861.0 \pm 782.2$ |
| StarLightZone Act3 | $2421.9 \pm 110.8$ | $2680.3 \pm 366.2$ |
| *Aggregate* | $2748.6 \pm 102.2$ | $3706.3 \pm 192.7$ |

Table 3: Detailed evaluation results for JERK.

| State | Score | Final Score |
|---:|---|---|
| AngelIslandZone Act2 | $1305.2 \pm 13.3$ | $1605.1 \pm 158.7$ |
| CasinoNightZone Act2 | $2231.0 \pm 556.8$ | $2639.7 \pm 799.5$ |
| FlyingBatteryZone Act2 | $1384.9 \pm 13.0$ | $1421.8 \pm 25.0$ |
| GreenHillZone Act2 | $3702.1 \pm 199.1$ | $4862.2 \pm 178.7$ |
| HillTopZone Act2 | $1901.6 \pm 56.0$ | $1840.4 \pm 326.8$ |
| HydrocityZone Act1 | $2613.0 \pm 149.6$ | $3895.5 \pm 50.0$ |
| LavaReefZone Act1 | $267.1 \pm 71.6$ | $200.3 \pm 71.9$ |
| MetropolisZone Act3 | $2623.7 \pm 209.2$ | $3291.4 \pm 398.2$ |
| ScrapBrainZone Act1 | $1442.6 \pm 108.8$ | $1756.3 \pm 314.2$ |
| SpringYardZone Act1 | $838.9 \pm 186.1$ | $829.2 \pm 158.2$ |
| StarLightZone Act3 | $2633.5 \pm 23.4$ | $3033.3 \pm 53.8$ |
| *Aggregate* | $1904.0 \pm 21.9$ | $2306.8 \pm 74.0$ |

Table 4: Detailed evaluation results for PPO.

| State | Score | Final Score |
|---:|---|---|
| AngelIslandZone Act2 | $1491.3 \pm 537.8$ | $2298.3 \pm 1355.8$ |
| CasinoNightZone Act2 | $2517.8 \pm 1033.0$ | $2343.6 \pm 1044.5$ |
| FlyingBatteryZone Act2 | $1105.8 \pm 177.3$ | $1305.7 \pm 221.9$ |
| GreenHillZone Act2 | $2477.6 \pm 435.3$ | $2655.7 \pm 373.4$ |
| HillTopZone Act2 | $2408.0 \pm 140.4$ | $3173.1 \pm 549.7$ |
| HydrocityZone Act1 | $622.8 \pm 288.6$ | $433.5 \pm 348.4$ |
| LavaReefZone Act1 | $885.8 \pm 125.6$ | $683.9 \pm 206.3$ |
| MetropolisZone Act3 | $1007.6 \pm 145.1$ | $1058.6 \pm 400.4$ |
| ScrapBrainZone Act1 | $1162.0 \pm 202.8$ | $2190.8 \pm 667.5$ |
| SpringYardZone Act1 | $564.2 \pm 195.6$ | $644.2 \pm 337.4$ |
| StarLightZone Act3 | $2134.4 \pm 313.4$ | $2519.0 \pm 98.8$ |
| *Aggregate* | $1488.8 \pm 42.8$ | $1755.1 \pm 65.2$ |

Table 5: Detailed evaluation results for Joint PPO.

| State | Score | Final Score |
|---:|---|---|
| AngelIslandZone Act2 | $3283.0 \pm 681.0$ | $4375.3 \pm 1132.8$ |
| CasinoNightZone Act2 | $5410.2 \pm 635.6$ | $6142.4 \pm 1098.7$ |
| FlyingBatteryZone Act2 | $1513.3 \pm 48.3$ | $1748.0 \pm 15.1$ |
| GreenHillZone Act2 | $8769.3 \pm 308.8$ | $8921.2 \pm 59.5$ |
| HillTopZone Act2 | $4289.9 \pm 334.2$ | $4688.6 \pm 109.4$ |
| HydrocityZone Act1 | $1249.8 \pm 206.3$ | $2821.7 \pm 154.1$ |
| LavaReefZone Act1 | $2409.0 \pm 253.5$ | $3076.0 \pm 13.7$ |
| MetropolisZone Act3 | $1409.5 \pm 72.9$ | $2004.3 \pm 110.4$ |
| ScrapBrainZone Act1 | $1634.6 \pm 287.0$ | $2112.0 \pm 713.9$ |
| SpringYardZone Act1 | $2992.9 \pm 350.0$ | $4663.4 \pm 799.5$ |
| StarLightZone Act3 | $1445.3 \pm 110.5$ | $2636.7 \pm 103.3$ |
| *Aggregate* | $3127.9 \pm 116.9$ | $3926.3 \pm 78.1$ |

Table 6: Detailed evaluation results for Joint Rainbow.

| State | Score | Final Score |
|---:|---|---|
| AngelIslandZone Act2 | $3770.5 \pm 231.8$ | $4615.1 \pm 1082.5$ |
| CasinoNightZone Act2 | $7877.7 \pm 556.0$ | $8851.2 \pm 305.4$ |
| FlyingBatteryZone Act2 | $2110.2 \pm 114.4$ | $2585.7 \pm 131.1$ |
| GreenHillZone Act2 | $6106.8 \pm 667.1$ | $6793.5 \pm 643.6$ |
| HillTopZone Act2 | $2378.4 \pm 92.5$ | $3531.3 \pm 4.9$ |
| HydrocityZone Act1 | $865.0 \pm 1.3$ | $867.2 \pm 0.0$ |
| LavaReefZone Act1 | $2753.6 \pm 192.8$ | $2959.7 \pm 134.1$ |
| MetropolisZone Act3 | $1340.6 \pm 224.0$ | $1843.2 \pm 253.0$ |
| ScrapBrainZone Act1 | $983.5 \pm 34.3$ | $2075.0 \pm 568.3$ |
| SpringYardZone Act1 | $2661.0 \pm 293.6$ | $4090.1 \pm 700.2$ |
| StarLightZone Act3 | $1813.7 \pm 94.5$ | $2533.8 \pm 239.0$ |
| *Aggregate* | $2969.2 \pm 170.2$ | $3704.2 \pm 151.1$ |

Table 7: Detailed evaluation results for humans.

| State | Score |
|---:|---|
| AngelIslandZone Act2 | $8758.3 \pm 477.9$ |
| CasinoNightZone Act2 | $8662.3 \pm 1402.6$ |
| FlyingBatteryZone Act2 | $6021.6 \pm 1006.7$ |
| GreenHillZone Act2 | $8166.1 \pm 614.0$ |
| HillTopZone Act2 | $8600.9 \pm 772.1$ |
| HydrocityZone Act1 | $7146.0 \pm 1555.1$ |
| LavaReefZone Act1 | $6705.6 \pm 742.4$ |
| MetropolisZone Act3 | $6004.8 \pm 440.4$ |
| ScrapBrainZone Act1 | $6413.8 \pm 922.2$ |
| SpringYardZone Act1 | $6744.0 \pm 1172.0$ |
| StarLightZone Act3 | $8597.2 \pm 729.5$ |
| *Aggregate* | $7438.2 \pm 624.2$ |

# D   Scores on Training Set

Table 8: Final performance for the joint Rainbow model over the last 10 episodes for each environment. Error margins are computed using the standard deviation over three runs.

| State | Score | State | Score |
| --- | --- | --- | --- |
| AngelIslandZone Act1 | $4765.6 \pm 1326.2$ | LaunchBaseZone Act2 | $1850.1 \pm 124.3$ |
| AquaticRuinZone Act1 | $5382.3 \pm 1553.1$ | LavaReefZone Act2 | $820.3 \pm 80.9$ |
| AquaticRuinZone Act2 | $4752.7 \pm 1815.0$ | MarbleGardenZone Act1 | $2733.2 \pm 232.1$ |
| CarnivalNightZone Act1 | $3554.8 \pm 379.6$ | MarbleGardenZone Act2 | $180.7 \pm 150.2$ |
| CarnivalNightZone Act2 | $2613.7 \pm 46.4$ | MarbleZone Act1 | $4127.0 \pm 375.9$ |
| CasinoNightZone Act1 | $2165.7 \pm 75.9$ | MarbleZone Act2 | $1615.7 \pm 47.6$ |
| ChemicalPlantZone Act1 | $4483.5 \pm 954.6$ | MarbleZone Act3 | $1595.1 \pm 77.6$ |
| ChemicalPlantZone Act2 | $2840.4 \pm 216.4$ | MetropolisZone Act1 | $388.9 \pm 184.2$ |
| DeathEggZone Act1 | $2334.3 \pm 61.0$ | MetropolisZone Act2 | $3048.6 \pm 1599.9$ |
| DeathEggZone Act2 | $3197.8 \pm 32.0$ | MushroomHillZone Act1 | $2076.0 \pm 1107.8$ |
| EmeraldHillZone Act1 | $9273.4 \pm 385.8$ | MushroomHillZone Act2 | $2869.1 \pm 1150.4$ |
| EmeraldHillZone Act2 | $9410.1 \pm 421.1$ | MysticCaveZone Act1 | $1606.8 \pm 776.9$ |
| FlyingBatteryZone Act1 | $711.8 \pm 99.1$ | MysticCaveZone Act2 | $4359.4 \pm 547.5$ |
| GreenHillZone Act1 | $4164.7 \pm 311.2$ | OilOceanZone Act1 | $1998.8 \pm 10.0$ |
| GreenHillZone Act3 | $5481.3 \pm 1095.1$ | OilOceanZone Act2 | $3613.7 \pm 1244.9$ |
| HiddenPalaceZone | $9308.9 \pm 119.1$ | SandopolisZone Act1 | $1475.3 \pm 205.1$ |
| HillTopZone Act1 | $778.0 \pm 8.1$ | SandopolisZone Act2 | $539.9 \pm 0.7$ |
| HydrocityZone Act2 | $825.7 \pm 2.2$ | ScrapBrainZone Act2 | $692.6 \pm 67.6$ |
| IcecapZone Act1 | $5507.0 \pm 167.5$ | SpringYardZone Act2 | $3162.3 \pm 38.7$ |
| IcecapZone Act2 | $3198.2 \pm 774.7$ | SpringYardZone Act3 | $2029.6 \pm 211.3$ |
| LabyrinthZone Act1 | $3005.3 \pm 197.8$ | StarLightZone Act1 | $4558.9 \pm 1094.1$ |
| LabyrinthZone Act2 | $1420.8 \pm 533.0$ | StarLightZone Act2 | $7105.5 \pm 404.2$ |
| LabyrinthZone Act3 | $1458.7 \pm 255.4$ | WingFortressZone | $3004.6 \pm 7.1$ |
| LaunchBaseZone Act1 | $2044.5 \pm 601.7$ | *Aggregate* | $3151.7 \pm 218.2$ |

Table 9: Final performance for the joint PPO model over the last 10 episodes for each environment. Error margins are computed using the standard deviation over two runs.

| State | Score | State | Score |
|---|---|---|---|
| AngelIslandZone Act1 | $9668.2 \pm 117.0$ | LaunchBaseZone Act2 | $1836.0 \pm 545.0$ |
| AquaticRuinZone Act1 | $9879.8 \pm 4.0$ | LavaReefZone Act2 | $2155.1 \pm 1595.2$ |
| AquaticRuinZone Act2 | $8676.0 \pm 1183.2$ | MarbleGardenZone Act1 | $3760.0 \pm 108.5$ |
| CarnivalNightZone Act1 | $4429.5 \pm 452.0$ | MarbleGardenZone Act2 | $1366.4 \pm 23.5$ |
| CarnivalNightZone Act2 | $2688.2 \pm 110.4$ | MarbleZone Act1 | $5007.8 \pm 172.5$ |
| CasinoNightZone Act1 | $9378.8 \pm 409.3$ | MarbleZone Act2 | $1620.6 \pm 30.9$ |
| ChemicalPlantZone Act1 | $9825.0 \pm 6.0$ | MarbleZone Act3 | $2054.4 \pm 60.8$ |
| ChemicalPlantZone Act2 | $2586.8 \pm 516.9$ | MetropolisZone Act1 | $1102.8 \pm 281.5$ |
| DeathEggZone Act1 | $3332.5 \pm 39.1$ | MetropolisZone Act2 | $6666.7 \pm 53.0$ |
| DeathEggZone Act2 | $3141.5 \pm 282.5$ | MushroomHillZone Act1 | $3210.2 \pm 2.7$ |
| EmeraldHillZone Act1 | $9870.7 \pm 0.3$ | MushroomHillZone Act2 | $6549.6 \pm 1802.9$ |
| EmeraldHillZone Act2 | $9901.6 \pm 18.9$ | MysticCaveZone Act1 | $6755.9 \pm 47.8$ |
| FlyingBatteryZone Act1 | $1642.4 \pm 512.9$ | MysticCaveZone Act2 | $6189.6 \pm 16.6$ |
| GreenHillZone Act1 | $7116.0 \pm 2783.5$ | OilOceanZone Act1 | $4938.8 \pm 13.3$ |
| GreenHillZone Act3 | $9878.5 \pm 5.1$ | OilOceanZone Act2 | $6964.9 \pm 1929.3$ |
| HiddenPalaceZone | $9918.3 \pm 1.4$ | SandopolisZone Act1 | $2548.1 \pm 80.8$ |
| HillTopZone Act1 | $4074.2 \pm 370.1$ | SandopolisZone Act2 | $1087.5 \pm 21.5$ |
| HydrocityZone Act2 | $4756.8 \pm 3382.3$ | ScrapBrainZone Act2 | $1403.7 \pm 3.3$ |
| IcecapZone Act1 | $5389.9 \pm 35.6$ | SpringYardZone Act2 | $9306.8 \pm 489.1$ |
| IcecapZone Act2 | $6819.4 \pm 67.9$ | SpringYardZone Act3 | $2608.1 \pm 113.2$ |
| LabyrinthZone Act1 | $5041.4 \pm 194.6$ | StarLightZone Act1 | $6363.6 \pm 198.7$ |
| LabyrinthZone Act2 | $1337.9 \pm 61.9$ | StarLightZone Act2 | $8336.1 \pm 998.3$ |
| LabyrinthZone Act3 | $1918.7 \pm 33.5$ | WingFortressZone | $3109.2 \pm 50.9$ |
| LaunchBaseZone Act1 | $2714.0 \pm 17.7$ | *Aggregate* | $5083.6 \pm 91.8$ |